\documentclass[mathpazo]{csam}
\usepackage{cite}
\usepackage{amsmath}
\usepackage{mathtools}
\usepackage{amssymb}
\usepackage{tabu}
\usepackage{tabularx}
\usepackage{booktabs}
\usepackage{algorithm}
\usepackage{algorithmic}
\usepackage{array}
\usepackage{multirow,multicol}
\usepackage[pdftex]{graphicx}

\usepackage[caption=false,font=footnotesize]{subfig}
\graphicspath{./figs/}
\begin{document}
\title{Hartley Spectral Pooling for Deep Learning}

\author[Hao Zhang et~al.]{Hao Zhang\textsuperscript{1}, Jianwei Ma\textsuperscript{1,2}\corrauth}
\address{1. Department of Mathematics and Artificial Intelligence Laboratory, Harbin Institute of Technology, Harbin 150001, P.R. China. \\
2. School of earth and space sciences, Peking University, Beijing 100000, P.R. China}
\emails{{\tt hao.zhang.hit@stu.hit.edu.cn, jma@pku.edu.cn} (J. Ma)}


\begin{abstract}
In most convolution neural networks (CNNs), downsampling hidden layers is adopted for increasing computation efficiency and the receptive field size. Such operation is commonly called pooling. Maximation and averaging over sliding windows (\emph{max/average pooling}), and plain downsampling in the form of strided convolution are popular pooling methods. Since the pooling is a lossy procedure, a motivation of our work is to design a new pooling approach for less lossy in the dimensionality reduction. Inspired by the spectral pooling proposed by Rippel et.al.\cite{rippel2015spectral}, we present the Hartley transform based spectral pooling method. The proposed spectral pooling avoids the use of complex arithmetic for frequency representation, in comparison with Fourier pooling. The new approach preserves more structure features for network's discriminability than max and average pooling. We empirically show the Hartley pooling gives rise to the convergence of training CNNs on  MNIST and CIFAR-10 datasets.
\end{abstract}

\ams{68T07
}
\keywords{Hartley transform, spectral pooling, deep learning.}

\maketitle

\section{Introduction}
\label{sec1}
Convolutional neural networks(CNNs) \cite{lecun1998gradient, schmidhuber2015deep, lecun2015deep} have been dominant machine learning approach for computer vision, and have spreaded out in many other fields. The modern framework of CNNs was established by LeCun et.al.\cite{lecun1990handwritten} in 1990, with three main components: convolution, pooling, and activation. Pooling is an important component of CNNs. Even before the resuscitation of CNNs, pooling was utilized to extract features to gain dimension-reduced feature vectors and acquire the invariance to small transformations of the input. This is motivated by the seminal work about complex cells in animal visual cortex by Hubel and Wiesel \cite{hubel1962receptive}. 

\begin{figure}[!t]
\centering
\includegraphics[width=5.6in]{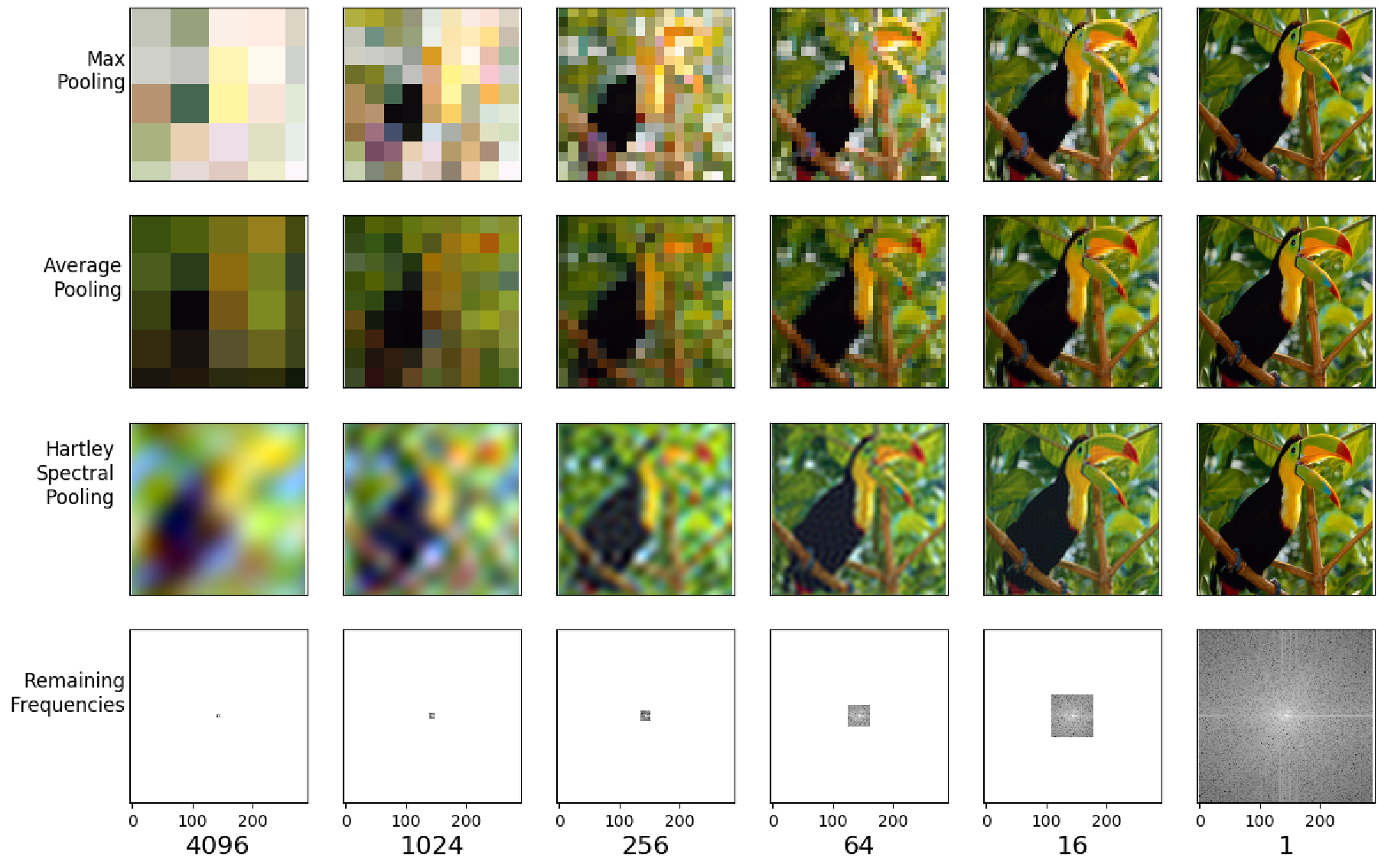}
\caption{Downsampling at different scales of dimensionality reduction. Hartley-based spectral pooling project real input onto the Hartley basis and truncates the real frequency representation as desired. This retains significantly more information as well as allows us choose arbitrary output dimension.}
\label{fig1}
\end{figure}

Pooling is of crucial for reducing computation cost, improving some amount of translation invariance and increasing the receptive field of neural networks. Numerous variants of pooling processes are proposed for classification accuracy improvement. These variants are mainly casted in four major categories based on value, rank, probability and transformed domain pooling methods, which are throughly reviewed recently in \cite{nadeem2020interpretation}. In shallow or mid-sized networks, max or average pooling are most widely used such as in AlexNet \cite{krizhevsky2012imagenet}, VGG \cite{simonyan2014very}, and GoogleNet \cite{szegedy2015going}. After Springenberg et al. \cite{springenberg2014striving} empirically revealed that strided convolution could replace pooling without loss of accuracy in classification task, deeper networks always use strided convolution for architecture-design simplicity. The most markable one of those exemplars is ResNet \cite{he2016deep}. However, most methods present a number of issues. For example, max pooling implies an amazing by-product of discarding at least 75\% of data. It can overfit the training data and does not guarantee generalization on test data. The maximum value picked out in each window only reflects very rough information. Average pooling, stretching to the opposite end, results in a gradual, constant attenuation of the contribution of individual grid in each window, and ignores the importance of local structure. 
These two poolings both sufer from sharp dimensionality reduction and lead to implausible looking results (see the first and second row in Figure \ref{fig1}). Strided convolution may cause aliasing since it simply picks one node in a fixed position in each local window \cite{proakis2013digital}, regardless of the significance of its activation.

There have been a few attempts to mitigate the harmful effects of max and average pooling, such as a linear combination and extension of them \cite{lee2016generalizing}, and nonlinear pooling layers \cite{gulcehre2014learned,bruna2013signal}. In most of the common implementations, max or average related pooling layers directly downscale the spatial dimension of feature maps by a factor. $L_p$ pooling \cite{bruna2013signal} provides better generalization than max pooling, with $p=1$ corresponding to average pooling and $p=\infty$ reducing to max pooling. Yu et al. \cite{yu2014mixed} proposed 
the mixed pooling, which combines max pooling and average pooling and switches between these two pooling methods randomly. Instead of picking the maximum values within each pooling region, stochastic pooling \cite{zeiler2013stochastic} and S3Pool \cite{zhai2017s3pool} stochastically pick a node in a window, and the former favors strong activations. In some networks, stride convolutions are also used for pooling. Notably, these pooling methods are all of integer stride larger than 1. To abate the loss of information caused by the dramatic dimension reduction, fractional max-pooling \cite{graham2014fractional} randomly generates pooling region with stride 1 or 2 to achieve pooling stride of less than 2. There are also some other pooling methods by applying/learning filters. For example, detail preserving pooling \cite{saeedan2018detail} uses inverse bilateral filter and learns two reward parameters in inverse bilateral weights from data to adaptively preserve the important details of feature maps. LEAP pooling \cite{sun2017learning} learns a shared linear filter over feature maps and aggregates the features within pooling region.
We refer these pooling methods mentioned above to as \emph{spatial pooling}. 

In 2015, Rippel et.al.\cite{rippel2015spectral} proposed the \emph{spectral pooling}, which downsamples the feature maps in frequency domain using low-pass filtering. It selects pooling region in Fourier based frequency domain by extracting low frequency subset. This approach can alleviates the issues of spatial pooling strategies as mentioned above, and it shows good information preserving ability. However, it introduces the processing of imaginary, which should be carefully treated in real CNNs. Moreover, the truncation in frequency domain may destroy the \textit{conjugate symmetry} of the Fourier frequency representation of the real input. For remediation some extra operations should be added to make sure the downsampled spatial approximation be real. This would be tedious and sometimes computation/time consuming. Following the work of \cite{rippel2015spectral},  spectral pooling approach using discrete cosine transform (DCT)\cite{smith2018cosine} and wavelet transform \cite{travis2018wavelet} are also proposed.

Inspired by the work of \cite{rippel2015spectral}, we present the Hartley transform-based spectral pooling in this paper. Our presented approach avoids the use of complex arithmetic and it could be plugged in modern CNNs effortlessly. Moreover, we provide a useful observation that preserving more information could contribute to the convergence of training modern CNNs.

\section{Method}

\subsection{Hartley transform}
\label{hartley}
The Hartley transform is an integral transform closely related to the Fourier transform \cite{hartley1942more,bracewell1986hartley}. It has some advantages over the Fourier transform in the analysis of real signals as it avoids the use of complex arithmetic. These advantages attracts researchers to conduct plenty of researches on its application and fast implementation during the 1990s \cite{bracewell1983discrete,bracewell1984fast,millane1994analytic,agbinya1987fast}.

In two dimensions, the Hartley transform, $H(u_1, u_2)$, of $f(x_1, x_2)$ is defined by \cite{millane1994analytic}
\begin{equation}
\begin{split}
H(u_1, u_2) =&\int^{\infty}_{-\infty} \int^{\infty}_{-\infty} f(x_1,x_2)[cos(2\pi (u_1x_1+u_2x_2)) \\
		&+ sin(2\pi (u_1x_1+u_2x_2))]dx_1dx_2
\end{split}
\end{equation}
and the inverse transform by
\begin{equation}
\begin{split}
f(x_1,x_2) = &\int^{\infty}_{-\infty} H(u_1, u_2)[cos(2\pi (u_1x_1+u_2x_2)) \\
		&+ sin(2\pi (u_1x_1+u_2x_2))]du_1du_2
\end{split}
\end{equation}
We write the kernel in the Hartley transform as 
\begin{equation}
\begin{matrix}
cas(x) &=& cos(x) + sin(x) \\
	& = & \frac{1-i}{2}e^{ix} +  \frac{1+i}{2}e^{-ix}
\end{matrix}
\end{equation}
so that
\begin{equation}
e^{ix} = \frac{1+i}{2}cas(x) + \frac{1-i}{2}cas(-x)
\end{equation}
Then the relation between the Hartley and Fourier transform can be derived from the above expressions, giving
\begin{equation}
H(u_1, u_2) = \frac{1+i}{2}F(u_1,u_2) + \frac{1-i}{2}F(-u_1,-u_2)
\end{equation}
and
\begin{equation}
F(u_1,u_2) = \frac{1-i}{2}H(u_1,u_2) + \frac{1+i}{2}H(-u_1,-u_2)
\end{equation}
where $F(u_1,u_2)$ represents the Fourier transform of $f(x_1,x_2)$.
In the case that function $f(x_1,x_2)$ is real, its Fourier transform is Hermetian, i.e.
\begin{equation}
F(-u_1,-u_2) = F^*(u_1,u_2)
\end{equation}
so the Fourier transform processes some redundancy on the real $u_1$-$u_2$ plane, which results in conjugate-symmetry constriction aiming at reducing training parameters in the frequency domain neural networks \cite{rippel2015spectral,pratt2017fcnn}.

The Hermetian property above shows that the Hartley transform of a real function can be written as
\begin{equation}
\label{equ1}
\begin{matrix}
H(u_1,u_2) & = & \mathcal{R}\{F(u_1,u_2)\} - \mathcal{I}\{F(u_1,u_2)\} 
\end{matrix}
\end{equation}
where $\mathcal{R}\{\cdot\}$ and $\mathcal{I}\{\cdot\}$ denote the real and imaginary parts respectively.
Note that given the definition above, the Hartley transform $\mathcal{H}$ is a real linear operator. It is symmetric, an involution and thus a unitary operator
\begin{equation}
f = \mathcal{H}\{\mathcal{H}f\}
\end{equation}
In the case of Hartley transform, imaginary part and conjuate symmetry no more need concerns for real inputs such as images. Note, moreover, that this does not increase any storage.

\paragraph{Differentiation.} Here we discuss how to propagate the gradient through the Hartley transform, which will be used in CNNs.  Define $x\in \mathbb{R}^{M\times N}$ and $y = \mathcal{H}(x)$ to be the input and output of a discrete Hartlay transform (DHT) respectively, and $L : \mathbb{R}^{M\times N} \to \mathbb{R}$ a real-valued loss function applied to $y$. Since the DHT is a linear operator, its gradient is simply the transform matrix itself. By the unitarity of DHT, the gradient in back-propagation corresponds to applying Hartley transform :
\begin{equation}
\frac{\partial L}{\partial x} = \mathcal{H}(\frac{\partial L}{\partial y})
\end{equation}
\subsection{Hartley-based spectral pooling}
\label{spectral}
Spectral pooling preserves considerably more information and structures for the same number of parameters \cite{rippel2015spectral}, as shown in the third row of Figure \ref{fig1}. This is because spectral transform provides a sparse basis in frequency domain for the inputs that have spatial structures. The spectrum power of a typical input is heavily concentrated in lower frequencies while higher frequencies mainly tend to encode noise \cite{torralba2003statistics}. This non-uniformity of spectrum power enables the removal of high frequencies with minimal damage of input information. 

To avoid the extra computation for conjugate symmetry ensurance which may be computation and time consuming in \cite{rippel2015spectral}, we suggest the Hartley transform-based spectral pooling. This spectral pooling is straightforward to understand and much easier to implement. Assume we have an input $x\in \mathbb{R}^{H\times W}$, and some desired output map dimensionality $h\times w$. First, we compute the DHT of the input into the frequency domain as $y = \mathcal{H}(x)\in \mathbb{R}^{H\times W}$, and shift the DC component of the input to the center of the domain. 
Then we crop the frequency representation by maintaining only the central $h\times w$ submatrix of frequencies, denoted as $\hat{y} \in \mathbb{R}^{h\times w}$. Finally, we take the DHT again as $\hat{x} = \mathcal{H}(\hat{y})$ to map frequency approximation back into spatial domain, obtaining the downsampled spatial approximation. The back-propagation of this spectral pooling is similar to its forward-propagation since Hartley transform is differentiable.

Those steps in both forward and backward propagation of this spectral pooling are listed in Algorithm \ref{alg1} and \ref{alg2}, respectively. These algorithms simplify the spectral pooling by Fourier transform, profited from that the Hartley transform of a real function is real rather than complex. 
Figure \ref{fig1} demonstrates the effect of this spectral pooling for various dimensionality reduction factors.

\begin{algorithm}[H]
\caption{Hartley Spectral pooling}
\label{alg1}
  \begin{algorithmic}[1]	
	\REQUIRE Map $x\in \mathbb{R}^{H\times W}$, output size $h\times w$ 
	\ENSURE Pooled map $\hat{x} \in \mathbb{R}^{h\times w}$
	\STATE $y \leftarrow \mathcal{H}(x)$ 
	\STATE $\hat{y} \leftarrow CropSpectrum(y, h\times w)$
	\STATE $\hat{x} \leftarrow \mathcal{H}(\hat{y})$
  \end{algorithmic}
\end{algorithm}

\begin{algorithm}[H]
\caption{Hartley Spectral pooling back-propagation}
\label{alg2}
  \begin{algorithmic}[1]
	\REQUIRE  Gradients w.r.t. output $\frac{\partial L}{\partial \hat{x}}$
	\ENSURE Gradients w.r.t. input $\frac{\partial L}{\partial x}$
	\STATE $\hat{z} \leftarrow \mathcal{H}(\frac{\partial L}{\partial \hat{x}})$
	\STATE $z \leftarrow PadSpectrum(\hat{z}, H\times W)$
	\STATE $\frac{\partial L}{\partial x} \leftarrow \mathcal{H}(z)$
  \end{algorithmic}
\end{algorithm}

\section{Experiments}
\label{experiment}

We verify the effectiveness of the Hartley-based spectral pooling through image classification task on MNIST, Fashion-MNIST and CIFAR-10 datasets. The trained networks include a toy CNN model (Table \ref{table1}), ResNet-16 and ResNet-20 \cite{he2016deep}. The toy network uses max pooling while ResNets employs strided convolutions for downscaling. In these experiments, spectral pooling shows favorable results. We also compare our DHT spectral pooling method with the DCT spectral pooling method \cite{smith2018cosine}.
Our implementation is based on PyTorch \cite{paszke2017automatic}.

\subsection{Datasets and configureations}
\paragraph{MNIST}The MNIST database \cite{kussul2004improved} is a large database of handwritten digits that is commonly used for benchmarking various convolutional neural networks. This dataset contains 60000 training examples and 10000 testing examples. All these examples are gray images in size of $28\times28$. In our experiment, we do not perform any preprocessing or augmentation on this dataset. Adam optimization algorithm \cite{kingma2014adam} is used in all experiments of classification on MNIST, with hyper-parameter $\beta_1=0.9$ and $\beta_2=0.999$ configured as suggested and a mini-batch size of 100. The initial learning rate is set to 0.001 and is divided by 10 every 5 epochs. Regularization is aborted in these experiments.
\paragraph{Fashion-MNIST}The Fashion-MNIST\cite{xiao2017fashion} dataset shares the same image size, data format and the
structure of training and testing splits with MNIST. But it is a more challenging alternative dataset for benchmarking machine learning methods since MNIST dataset is too easy. If the methods do work on MNIST, they may still fail on others. In our experiments, we set the training configurations same as that in MNIST. 
\paragraph{CIFAR-10}The CIFAR-10 dataset consists of 60000 colored natural images in 10 classes, with 6000 images per class holding 5000 for training and 1000 for testing. Each image is in size of $32\times32$. For data augmentation we follow the practice in     \cite{he2016deep}, doing horizontal flips and randomly sampling $32\times32$ crops from image padded by 4 pixels on each side. The normalization is performed in data preprocessing by using the channel means and standard deviations. In experiments on this dataset, we use stochastic gradient method with Nesterov momentum and cross-entropy loss. The initial learning rate is set to 0.1, and is multiplied by 0.1 at 80 and 120 epochs. The weight decay parameter is configured to $10^{-4}$ and the momentum is set to 0.9 without dampening. Mini-batch size is set to 128.
\begin{figure}
\centering
\includegraphics[width=5in]{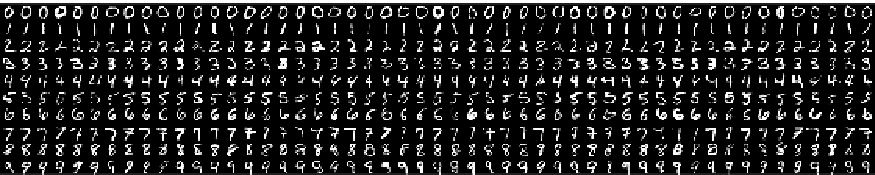}
\caption{\small{500 examples in MNIST dataset. From top row to bottom labels 0 to 9 respectively.}}
\end{figure}

\subsection{Classification results on MNIST and Fashion-MNIST}
\begin{figure}
\centering
	\begin{tabular}{@{}cc@{}}
	\includegraphics[width=2.6in]{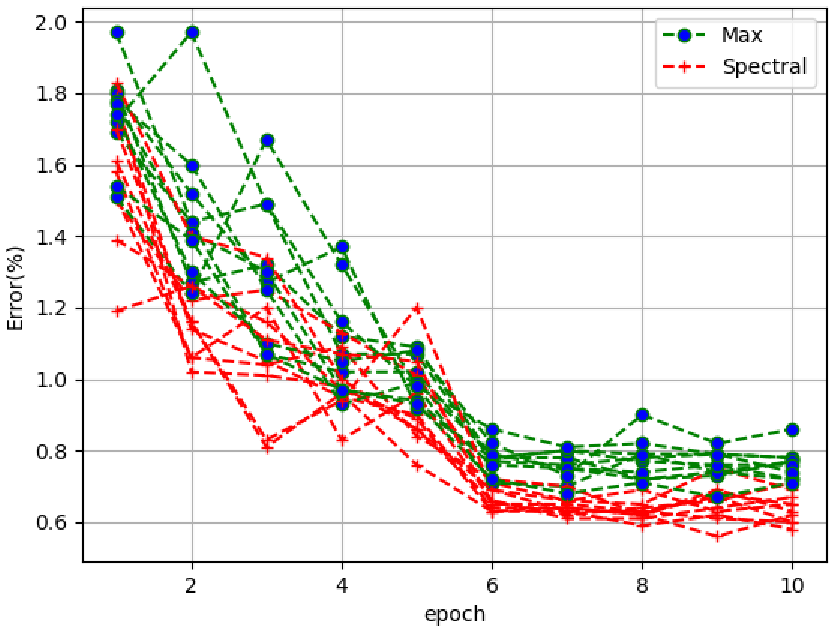} &
	\includegraphics[width=2.6in]{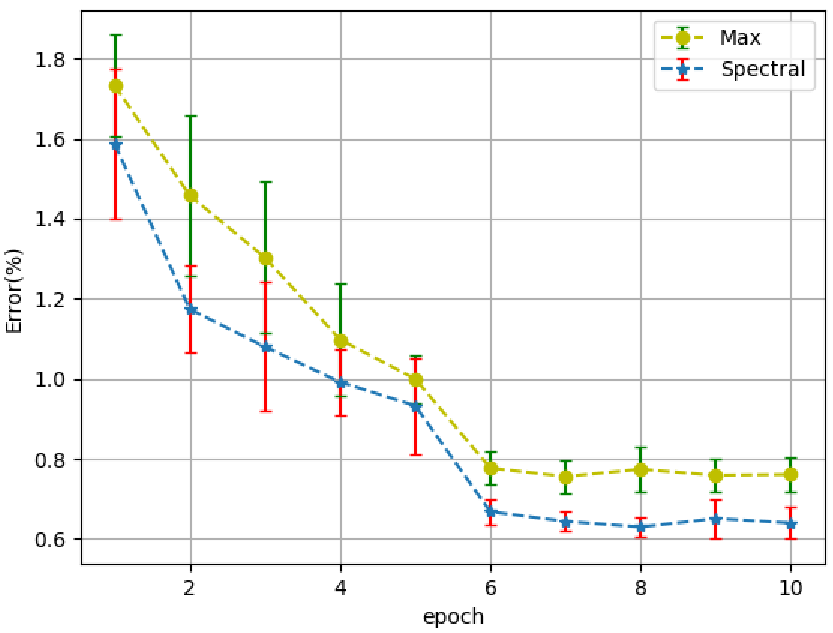} \\
	\end{tabular}
	\caption{Classification error on MNIST testing set by networks in Table \ref{table1}.
		(\textbf{Left}) Classification error curves in 10 runs. (\textbf{Right})
		\emph{mean$\pm$std} of ten runs, with  best error $0.605\%(0.63\pm 0.025)$, $0.719\%(0.759\pm 0.040)$ for 
		spectral pooling and max pooling respectively.}
\label{fig2}
\end{figure}

\paragraph{Shallow network}
We first conduct experiments using the toy network (see Table \ref{table1}) on MNIST. Each convolution layer is followed by a batch normalization and a ReLU nonlinearity. We test Hartley-based spectral pooling by replacing max pooling in this architecture. The training procedure lasts 10 epoches and is repeated 10 times. The classification error on testing set in each epoch is shown in Figure \ref{fig2}.

Compared to max pooling, spectral pooling shows strong results, yielding more than 15\% reduction on classification error observed in this experiment. As  all things equal except the pooling layers, we claim that this improvement is achieved by the better information-preserved ability of Hartley-based spectral pooling.

\begin{table}[htp]
\caption{The toy CNN model for classification on MNIST.}
\label{table1}
\centering
\begin{tabular}{c|c|c|c}
\toprule
layer name & output size  & Max Pooling model	&  Spectral Pooling model\\
\midrule
conv1		& 28$\times$28 & \multicolumn{2}{c}{5$\times$5, 16}	\\
\midrule
pool1 	& 14$\times$14 & Max, stride=2 		& Spectral, 14$\times$14 \\
\midrule
conv2		& 14$\times$14 & \multicolumn{2}{c}{5$\times$5, 32}	\\
\midrule
pool2 	& 7$\times$7 & Max, stride=2 		& Spectral, 7$\times$7 \\
\midrule
fc	& 1$\times$1 	& \multicolumn{2}{c}{10-d fc} 	\\
\bottomrule
\end{tabular}
\end{table}

\begin{figure}[!htpb]
\centering
\begin{tabular}{@{}cc@{}}
\includegraphics[width=3in]{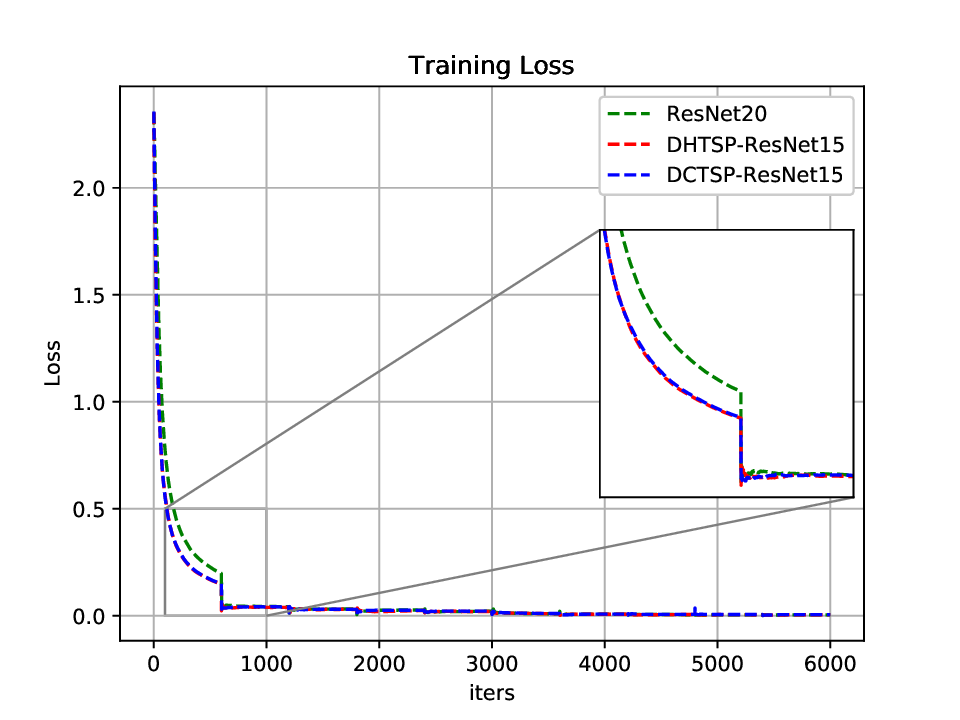} &
\includegraphics[width=3in]{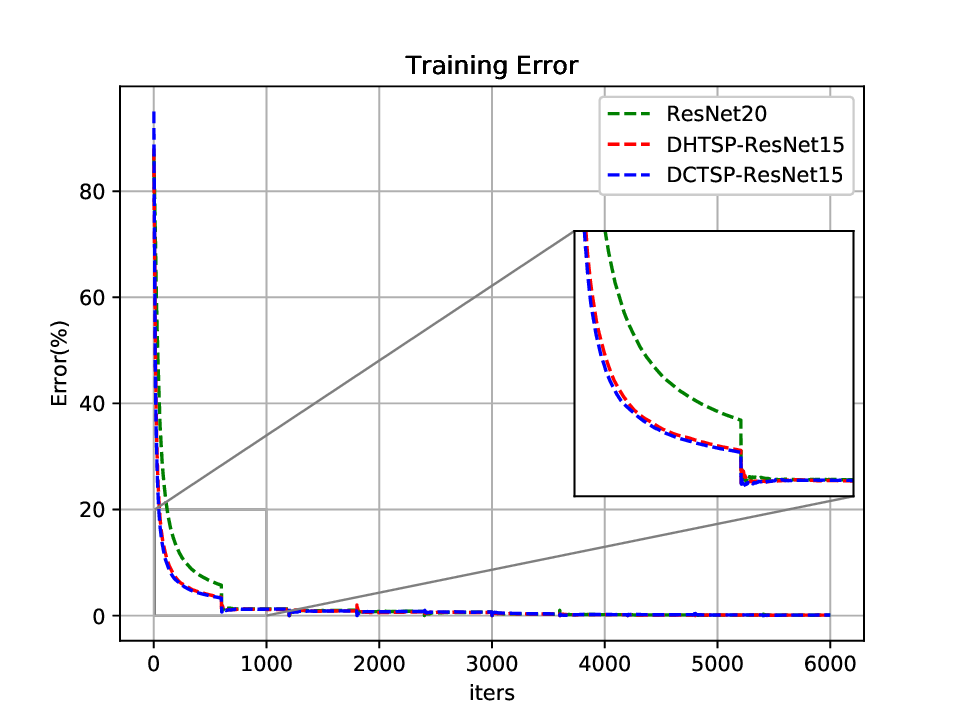} \\
\end{tabular}
\caption{Training on \textbf{MNIST} by ResNet20 and SP-ResNet15. (\textbf{Left}) training loss; (\textbf{Right}) classification error on training set.}
\label{fig3}
\end{figure}

\paragraph{ResNet}
Next, we conduct experiments on both MNIST and Fashion-MNIST datasets using ResNet-20 \cite{he2016deep}. We don't use more deeper residual net such as ResNet-110 because much more parameters in this architecture may give rise to overfitting. ResNet is composed of numerous residual building blocks. It is a much modern convolutional neural network which does not explicitly use pooling layers but instead embeds a stride-2 convolution layer inside some of building blocks for the 
implemenation of downsampling. 
In our experiments we replace the stride-2 convolutional layer by spectral pooling and remove the skip connection \cite{szegedy2015going,srivastava2015highway} in those downscaling blocks. Besides, we set the output size of serial spectral pooling layers linearly decreased (reducing 8 in each axis after a spectral pooling layer). The manually tuned network architecture is depicted in Table \ref{table2}. We leave the global average pooling untouched. Each run is repeated 5 times. The number of parameters in each network, the averaged  training time of single epoch, the best result as well as mean and standard deviation are reported in Table \ref{table3}. To distingush the different spectral pooling methods, we prefix SP-ResNet15 with the spectral transform name (DHT, DCT).

As shown in Table \ref{table3}, the DHTSP-ResNet15 obtains slightly better testing error ($+$0.04\%) than the plain ResNet20 in MNIST task, and a significant improvement ($+$0.65\%) in Fashion-MNIST task, even though it holds parameters nearly half of that in ResNet20. The training time increases 4.2 s for a single epoch with the replacement of strided convolution by DHT spectral pooling. The DCTSP-ResNet15 performs slightly better ($+$0.14\%) than our DHTSP-ResNet15 on Fashion-MNIST dataset, but the training time doubles. The slightly outperforming of the DCTSP over our DHTSP is not suprising because for DCT more energey is concentrated in even fewer spectra. Further, we illustrate one among the five training procedures in Figure \ref{fig3}. It is observed that the SP-ResNet15 converges faster than ResNet20 (left panel) and performs better in classification (right panel). This indicates the spectral pooling eases the optimization by providing faster convergence at the early stage.

\begin{table}[htpb]
\caption{The architecture of spectral pooling ResNet for MNIST and Fashion-MNIST. Building blocks are shown in brackets, with the numbers of block stacked. The SP stands for the spectral pooling layer, and the footnote $n\times n$ indicates the output size.}
\label{table2}
\centering
\begin{tabular}{c|c|c}
\toprule
block name & output size & SP-ResNet15\\
\midrule
conv1		&  28$\times$28 & 3$\times$3, 16 \\
\midrule
conv2   &  28$\times$28 &  $\begin{bmatrix}3\times3,16 \\ 3\times3,16\end{bmatrix}\times2$ \\
\midrule
downsample1	& 20$\times$20 & 	$\begin{bmatrix} SP_{20\times20} \\ 3\times3, 32 \end{bmatrix}$ \\
\midrule
conv3		& 20$\times$20	& $\begin{bmatrix}3\times3,32 \\ 3\times3,32\end{bmatrix}$ \\
\midrule
downsample2	& 12$\times$12 &  $\begin{bmatrix} SP_{12\times12} \\ 3\times3, 32 \end{bmatrix}$ \\
\midrule
conv4		& 8$\times$8 	& $\begin{bmatrix}3\times3,32 \\ 3\times3,32\end{bmatrix}$ \\
\midrule
downsample3	&  4$\times$4 &  $\begin{bmatrix} SP_{4\times4} \\ 3\times3, 64 \end{bmatrix}$ \\
\midrule
conv5	&  4$\times$4	& $\begin{bmatrix}3\times3,64 \\ 3\times3,64\end{bmatrix}$ \\
\midrule
	&	1$\times$1 	& avg pool, 10-d fc \\
\bottomrule
\end{tabular}
\end{table}



\begin{table}[htpb]
\caption{Classification error on \textbf{MNIST} (left)and \textbf{Fashion-MNIST}(right) testing set.  All methods are without data augmentation. The average training time of a single epoch is reported. For each method we run it 5 times and show "best(mean$\pm$std)".}
\label{table3}
\centering
\begin{tabular}{@{}c|c|c|c@{}}
\toprule
method		&  \#params &  training time (s)	& error(\%) \\
\hline
ResNet16 & 0.18M  &	25.2	& 0.36 (0.40$\pm$0.04)/ 7.14 (7.23$\pm$0.06)\\
ResNet20			& 0.27M  &	26.4	& 0.36  (0.40$\pm$0.04)/6.91(7.12$\pm$0.13) \\
DHTSP-ResNet15			& \multirow{2}{*}{\textbf{0.15M}} &  30.6	& \textbf{0.32}(0.37$\pm$0.04)/6.26 (6.38$\pm$ 0.07)\\
DCTSP-ResNet15			&  &  66.7 	& \textbf{0.32}(0.38$\pm$0.04)/\textbf{6.12} (6.21$\pm$0.07)\\

\bottomrule 
\end{tabular}
\end{table}

\subsection{Classification results on CIFAR-10}
For the experiments on CIFAR-10, the architecture of spectral pooling ResNet is similar to SP-ResNet15 that is used in MNIST case. We set the sizes of output of spectral pooling layers to be $24\times24$, $16\times16$, $8\times8$ sequentially. We use plain ResNet-16 as a counterpart, since it contains almost the same amount of parameters as SP-ResNet15 (see Table \ref{table4}). The DHTSP-ResNet15 outperforms  ResNet16 with smaller best testing error by 0.24\% and smaller mean testing error by 0.1\%, although the training time increases about 1000 s. The DCTSP-ResNet15 performs slightly worse than DHTSP-ResNet15, and doubles the training time. It gets even sligtly worse mean testing error than the plain ResNet-16.
\begin{table}[htpb]
\caption{Classification error on \textbf{CIFAR-10} testing set. All methods are with data augmentation. For each method we run it 5 times and show "best(mean$\pm$std)".}
\label{table4}
\centering
\begin{tabular}{@{}c|c|c|c@{}}
\toprule
method	& \# params & training time (s)	& error(\%) \\
\hline
ResNet16			& 0.18M 	& 2896.4	& 8.87 (9.06$\pm$0.10 )\\
DHTSP-ResNet15			& \multirow{2}{*}{\textbf{0.15M}} & 3900.8	& \textbf{8.63}(8.96$\pm$0.16) \\
DCTSP-ResNet15			&  &  7808.5	& 8.81(9.11$\pm$0.17) \\
\bottomrule 
\end{tabular}
\end{table}

\section{Conclusion}
We present a full real-valued Hartley spectral pooling method in this paper. Based on this approach, we provide some results on several commonly used benchmark dataset (MNIST, Fashion-MNIST and CIFAR-10) by training a toy CNN and the modified Resnets. We demonstrate the Hartley spectral pooling yields higher classification accuracy than its counterpart max pooling. We also investigate the contribution of this spectral pooling method to the convergence of training neural networks. In Resnet, it improves the results by expanding the space of spatial dimensionality of downsamplings. We also compare the Hartley spectral pooling with its spectral pooling counterpart, the discrete cosine transform spectral pooling. Although sometimes the Hartley spectral pooling perform slightly worse than the discrete cosine transform spectral pooling in respect of classfification accuracy, it is much faster. Moreover, the Hartley transform obeys an analogous convolution theorem, which means it has the potential for constructing full real spectral neural nets.

\section*{Acknowledgments}
The work is supported in part by National Key Research and Development Program of China under Grant 2017YFB0202902, and NSFC under Grant 41625017.


\end{document}